%% file: bmvc_final.tex
\documentclass{bmvc2k}

\usepackage{amsmath}
\usepackage{amssymb}
\usepackage{booktabs}
\usepackage{algorithm}
\usepackage{algpseudocode}
\usepackage{tikz}
\usetikzlibrary{positioning, arrows.meta, backgrounds, fit, shadows, calc}
\definecolor{cFrost}{HTML}{4E7FB0}   
\definecolor{cCTPR}{HTML}{2A9D8F}    
\definecolor{cTCR}{HTML}{3D5A80}     
\definecolor{cTRACE}{HTML}{E07A1F}   
\definecolor{cPanel}{HTML}{F3F6FB}   
\definecolor{cInk}{HTML}{262626}     
\tikzset{
  snowflake/.pic={
    \foreach \a in {0,60,120,180,240,300}{
      \begin{scope}[rotate=\a]
        \draw[line width=0.35pt, cFrost] (0,0) -- (0,1.7mm);
        \draw[line width=0.35pt, cFrost] (0,1.1mm) -- (0.55mm,1.55mm);
        \draw[line width=0.35pt, cFrost] (0,1.1mm) -- (-0.55mm,1.55mm);
      \end{scope}
    }
  }
}

\title{Closing the Estimation Gap: Training-Free Transductive Inference for Cross-Domain Few-Shot Action Recognition}

\addauthor{Changyi Li}{changyi.li@aalto.fi}{1}
\addauthor{Zeyu Yang}{zeyu.yang@aalto.fi}{1}
\addauthor{Yu Xiao}{yu.xiao@aalto.fi}{1}
\addinstitution{
 Dept.\ of Information and\\
 Communications Engineering\\
 Aalto University\\
 02150 Espoo, Finland
}

\runninghead{Li, Yang, Xiao}{TRACE for CD-FSAR}

\def\etal{\emph{et al}\bmvaOneDot}

\newcommand{\R}{\mathbb{R}}
\newcommand{\Sset}{\mathcal{S}}
\newcommand{\Qset}{\mathcal{Q}}
\newcommand{\sm}{\mathrm{softmax}}
\newcommand{\mean}{\mathrm{mean}}
\newcommand{\sotd}{\textsc{Trace}}      
\newcommand{\bH}{\widehat{H}}

\begin{document}

\maketitle

\begin{abstract}
Cross-domain few-shot action recognition (CD-FSAR) aims to recognise unseen action classes in a target domain from a handful of labelled videos. Despite advances in representation learning, CD-FSAR inference still relies on prototypes estimated from extremely limited support examples. To assess this bottleneck, we perform an oracle analysis with the feature representation fixed. Constructing oracle prototypes and decision boundaries from ground-truth query labels improves 5-way 1-shot accuracy by 7--20 percentage points across five target datasets. This reveals a substantial \emph{estimation gap}: the representation contains useful discriminative information, but prototypes and boundaries estimated from scarce supports fail to capture each episode's true class structure. Motivated by this finding, we propose \sotd{}, a training-free transductive inference framework that improves episode-level estimation without modifying the backbone. \sotd{} comprises two complementary stages. Confidence-Thresholded Transductive Prototype Refinement (CTPR) refines support-based prototypes using confidently predicted queries. The refined prototypes initialise a lightweight per-episode classifier, which Transductive Classifier Refinement (TCR) further adapts by encouraging confident, class-balanced predictions over all queries. Together, the two stages improve prototype and boundary estimation, enabling better use of information already present in the unlabelled queries. Across five target datasets (HMDB51, UCF101, SSv2, Diving48, and RareAct) in both 1-shot and 5-shot settings, \sotd{} consistently improves a strong CD-FSAR baseline, including gains of more than six percentage points on HMDB51 1-shot, while recovering a substantial fraction of the oracle-estimated gap. These results suggest that improving test-time estimation is a powerful, broadly applicable complement to representation learning for CD-FSAR under extreme support scarcity. Code: https://github.com/hisalaheiyo/TRACE.
\end{abstract}

\begin{figure}
\centering
\includegraphics[width=0.57\linewidth]{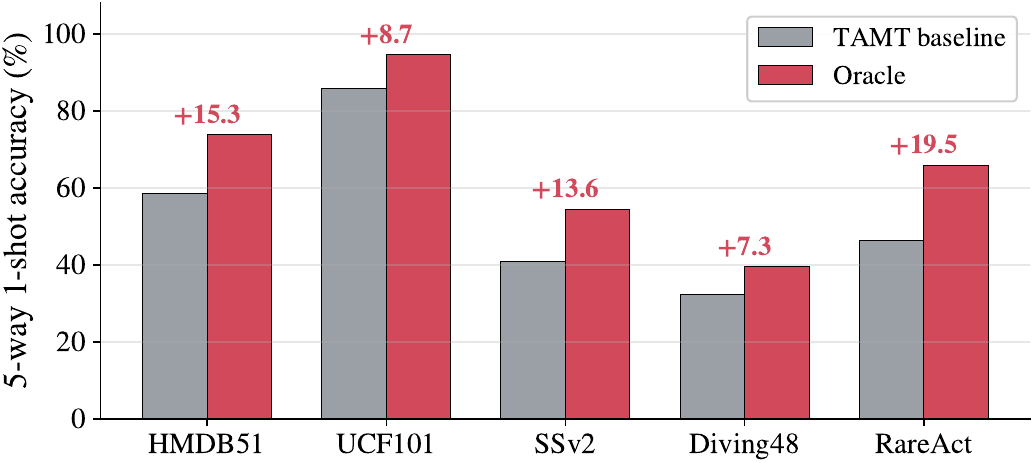}
\caption{\textbf{Oracle estimation reveals a substantial gap across target datasets (1-shot).}
Left bars: TAMT baseline; right bars: oracle estimation, both under the held-out-half protocol. Numbers above bars give the gap in percentage points.}
\label{fig:gap}
\end{figure}

\section{Introduction}
\label{sec:intro}

Large pre-trained video models have substantially advanced action recognition capabilities, but their success typically relies on abundant labelled data for every class of interest. Many real-world applications violate this assumption. Surveillance of rare events, sports analytics, clinical video analysis, and assistive monitoring~\cite{li2026reachability} often involve domain-specific actions that are expensive to annotate and differ substantially from the web-scale videos used for pre-training. Cross-domain few-shot action recognition (CD-FSAR) formalises this challenge: a model trained on a source domain must recognise unseen action classes in a different target domain using only a handful of labelled videos per class. The combination of severe domain shift and extreme label scarcity makes CD-FSAR considerably more challenging than either few-shot recognition or domain adaptation alone.

Most existing approaches address this challenge through representation learning. Recent methods design temporally-aware backbones, introduce lightweight adaptation modules, or develop specialised meta-training objectives to improve transferability across domains~\cite{tamt2025,cdfslv2023,molo2023}. While these advances are valuable, they largely operate before a target episode is observed. Once training is complete, inference typically reduces to matching each query video against prototypes estimated from only a few support examples. This raises a simple but largely unexplored question: given a strong and fixed representation, how much performance is still limited by episode-level estimation at inference time? By estimation we mean the construction of prototypes and decision boundaries from each episode's few labelled supports.

To answer this question, we perform an oracle analysis that isolates representation quality from estimation quality. Specifically, we keep the feature representation (the frozen encoder and its embeddings) fixed and construct oracle prototypes and decision boundaries using the ground-truth labels of the query videos within each episode. The resulting gains are large and consistent. Across five target datasets, oracle estimation improves 1-shot accuracy by 7--20 percentage points, as shown in Fig.~\ref{fig:gap}; on HMDB51, for example, accuracy increases from $58.4\%$ to $73.8\%$ under the held-out-half evaluation. Because the encoder remains unchanged, these improvements cannot be attributed to better representations. Instead, they reveal a substantial \emph{estimation gap}: under extreme support scarcity, prototypes and decision boundaries estimated from only a few support examples fail to capture the true class structure of an episode. Importantly, the information required to improve these estimates is already available at test time in the distribution of the unlabelled query videos.

Guided by this diagnosis, we seek to close the estimation gap without modifying the representation itself. We propose \sotd{}, a training-free transductive inference framework that operates entirely within a single test episode. \sotd{} consists of two complementary stages that target the two components of the estimation gap. Confidence-Thresholded Transductive Prototype Refinement (CTPR) improves prototype estimation by incorporating confidently predicted query videos into the support-based prototypes. The refined prototypes are then used to initialise a lightweight per-episode classifier. Transductive Classifier Refinement (TCR) subsequently improves boundary estimation by adapting this classifier using the information-maximisation principle of Boudiaf~\etal~\cite{boudiaf2020}, encouraging predictions over the query set to become both confident and class-balanced. Together, the two stages enable more accurate episode-level estimation while leaving the backbone entirely unchanged.

We evaluate \sotd{} on five target datasets (HMDB51, UCF101, SSv2, Diving48, and RareAct) under both 1-shot and 5-shot settings. A central question raised by our oracle analysis is how much of the estimation gap can be recovered without changing the representation. Across all five datasets, \sotd{} consistently recovers a substantial fraction of this gap, improving a strong CD-FSAR baseline by more than six percentage points on HMDB51 1-shot. Despite requiring no additional training, it remains competitive with state-of-the-art methods that modify the training procedure. Because \sotd{} only re-estimates episode-specific prototypes and decision boundaries, its benefits are largely independent of the underlying recognition framework. We verify this by applying \sotd{} unchanged to three different episodic heads, where it achieves the best average performance in all cases.

In summary, our contributions are three-fold:
\begin{itemize}
\item We reveal a \textbf{substantial estimation gap} in CD-FSAR under extreme support scarcity through an oracle diagnostic framework, and demonstrate that a large portion of this gap can be recovered at test time (Sec.~\ref{sec:analysis}).
\item We propose \textbf{\sotd{}}, a training-free transductive inference framework that improves both prototype estimation and boundary estimation through the combination of CTPR and TCR, without modifying the backbone or requiring additional training (Sec.~\ref{sec:method}).
\item We conduct \textbf{extensive experiments} across five datasets, two shot settings, and three episodic heads, demonstrating consistent improvements over a strong baseline and analysing when and why the two transductive estimators are complementary (Sec.~\ref{sec:exp}).
\end{itemize}

\section{Related Work}
\label{sec:related}

\paragraph{Cross-domain few-shot action recognition.}
Cross-domain few-shot action recognition (CD-FSAR) builds upon a large body of work on within-domain few-shot action recognition, including temporal alignment, relation-based matching, and transformer-based approaches such as OTAM~\cite{otam2020}, TRX~\cite{trx2021}, HyRSM~\cite{hyrsm2022}, STRM~\cite{strm2022}, and MoLo~\cite{molo2023}. CD-FSAR further introduces a substantial distribution shift between a label-rich source domain and a label-scarce target domain. Existing methods primarily address this challenge through representation learning or training-time adaptation. Representative approaches exploit self-training, distillation, self-supervised temporal learning, mixed-source distillation, or curriculum learning to improve cross-domain transfer~\cite{startup2021,dd2021,seen2023,dmsd2024,cdfslv2023}. Most relevant to our setting, TAMT keeps a frozen VideoMAE backbone~\cite{videomae2022} and adapts lightweight temporal modules through meta-training, achieving strong performance without joint source--target training~\cite{tamt2025}. Despite their differences, these methods share a common characteristic: cross-domain transfer is improved primarily through stronger representations or better training procedures before a target episode is observed. In contrast, we study a complementary question: once a strong representation is fixed, how much performance is still limited by episode-level estimation at inference time?

\paragraph{Transductive and test-time estimation in few-shot learning.}
A complementary line of work improves few-shot recognition at inference time by exploiting relationships among the query samples. In image-domain few-shot learning, transductive inference has been explored through label propagation, manifold smoothing, Laplacian regularisation, and optimal transport~\cite{tpn2019,embprop2020,laplacianshot2020,ptmap2020, li2025revisiting}. Transductive information maximisation further adapts a per-episode classifier by encouraging predictions to be both confident and class-balanced~\cite{boudiaf2020}, while test-time adaptation more broadly updates models on unlabelled test data through objectives such as entropy minimisation~\cite{tent2021}. Related ideas have also been explored for within-domain few-shot video recognition~\cite{tnt2021,nguyen2022inductive}. Another closely related direction is prototype rectification, which reduces prototype bias by incorporating unlabelled queries through pseudo-labelling and prototype re-estimation~\cite{protorect2020}.

Collectively, these methods demonstrate that the unlabelled query set contains information that can substantially improve episode-level decisions. However, they are typically evaluated in image or single-domain settings and are rarely grounded in an explicit diagnosis of which estimation errors they address. In contrast, we begin with an oracle analysis that identifies a substantial estimation gap in CD-FSAR and separates representation quality from estimation quality. This diagnosis provides the basis for our training-free test-time inference framework.

\section{An Oracle Analysis of the Estimation Gap}
\label{sec:analysis}

We begin by asking what limits CD-FSAR accuracy once a strong representation is fixed. Our analysis shows that a large and recoverable portion of the error arises from episode-level estimation: prototypes and decision boundaries must be inferred from only a few labelled support videos. This diagnosis directly motivates the design of our method in Sec.~\ref{sec:method}.

\paragraph{Setup and notation.}
A CD-FSAR episode is a $K$-way $N$-shot task sampled from a target domain whose classes are unseen during training. It contains a labelled \emph{support} set $\Sset={(x_i,y_i)}*{i=1}^{KN}$ with $N$ videos per class, and an unlabelled \emph{query} set $\Qset={x_j}*{j=1}^{KM}$ whose labels are to be predicted. Following TAMT~\cite{tamt2025}, a frozen feature extractor maps each video to an embedding $z=\phi(x)\in\R^{d}$ and is \textbf{never updated at test time}. The standard baseline forms a per-class prototype
\begin{equation}
\label{eq:proto}
p_k = \frac{1}{N}\sum_{i:y_i=k} z_i,
\end{equation}
and predicts $\hat y_j=\arg\max_k \mathrm{sim}(z_j,p_k)$, where $\mathrm{sim}$ is the inner product for $N{=}1$ and the negative squared Euclidean distance for $N{>}1$, matching the baseline metric.

\begin{table}
\centering
\small
\begin{tabular}{lccccc}
\toprule
Dataset & Baseline & TUB-P & TUB-C & TUB-I & gap $\Delta$ \\
\midrule
HMDB51   & 58.45 & 71.24 & 71.05 & \textbf{73.79} & $+15.34$ \\
UCF101   & 85.90 & \textbf{94.63} & 94.33 & 94.18 & $+8.73$ \\
SSv2     & 40.87 & 54.42 & 50.64 & \textbf{54.47} & $+13.60$ \\
Diving48 & 32.21 & \textbf{39.47} & 35.98 & 39.07 & $+7.26$ \\
RareAct  & 46.40 & 63.68 & 64.13 & \textbf{65.91} & $+19.51$ \\
\bottomrule
\end{tabular}
\caption{\textbf{Oracle upper bounds on 5-way 1-shot CD-FSAR.}
Baseline denotes TAMT prototype matching evaluated on the held-out query half. TUB-P, TUB-C, and TUB-I denote oracle prototype estimation, oracle boundary estimation, and classifier refinement initialised from oracle prototypes, respectively. Each oracle uses the true labels of the other query half and is evaluated on the same held-out half as the baseline. Gap $\Delta$ is the difference between the best oracle and the baseline. Best oracle per row is shown in \textbf{bold}.}
\label{tab:oracle}
\end{table}

\paragraph{Oracle protocol.}
With the representation fixed, the episode-specific quantities are the prototypes in Eq.~\ref{eq:proto} and, more generally, the decision boundaries induced by them. Rather than asking how to improve these estimates, we first ask how much performance could be gained if they were estimated with access to perfect label information. To quantify this, we construct three transductive upper bounds that use ground-truth query labels for analysis only. The term is diagnostic rather than formal: each oracle scores one specific label-assisted estimator on the fixed features, not the best possible decision rule. To avoid fitting the evaluation set, each oracle uses the labels of one half of the query set and is evaluated on the other half. The baseline is evaluated on the same held-out query half, so all performance gaps are directly comparable.

The \emph{Prototype Oracle} (TUB-P) recomputes prototypes from the support set and the true-labelled query half, estimating the best performance achievable by a prototype classifier with oracle prototype estimation. The \emph{Classifier Oracle} (TUB-C) fits a linear classifier on the support set and the true-labelled query half, estimating the benefit of oracle boundary estimation. The \emph{Initialisation Oracle} (TUB-I) optimises the same per-episode classifier as TUB-C but initialises it from the oracle prototypes, estimating the value of accurate prototype initialisation for boundary refinement.

\paragraph{A substantial estimation gap exists.}
Table~\ref{tab:oracle} compares the baseline with the three oracle upper bounds on all five target datasets in the 5-way 1-shot setting. Across all datasets, the oracle upper bounds consistently outperform the baseline by a large margin, indicating that support-based estimation leaves substantial room for improvement even when the representation is fixed. With the encoder and embeddings kept fixed, the best oracle improves accuracy by $+7.3$ (Diving48) to $+19.5$ (RareAct) percentage points; for example, HMDB51 increases from $58.4$ to $73.8$. Since the representation is unchanged throughout this comparison, the observed improvement arises from better episode-level estimation rather than from changes to the features themselves. These gains show that the fixed representation retains class information that support-based estimation leaves unused. Whether that information can be recovered from the unlabelled queries alone is an empirical question; Secs.~\ref{sec:method} and~\ref{sec:exp} show that much of it can.

Fig.~\ref{fig:gap} visualises this estimation gap across target datasets. The left bar shows the TAMT baseline, which estimates prototypes solely from the support set, while the right bar shows the best oracle under the same held-out-half evaluation protocol. Numbers above the bars indicate the gap in percentage points. The consistent separation between the two bars highlights that support-based episode-level estimation remains a substantial source of error under extreme label scarcity.

\paragraph{Prototype estimation and boundary estimation are complementary.}
Comparing the three oracles reveals how prototype estimation and boundary estimation interact. A freely fitted classifier (TUB-C) does not always outperform the Prototype Oracle. On more challenging temporal datasets, it underperforms TUB-P (SSv2: $50.6$ vs.\ $54.4$; Diving48: $36.0$ vs.\ $39.5$), suggesting that an unconstrained decision boundary can overfit when the labelled set is extremely small.

In contrast, the Initialisation Oracle (TUB-I) is the best or tied-best oracle on three of five datasets and remains within $0.5$ percentage points of the strongest oracle elsewhere. Because TUB-I optimises the same classifier as TUB-C but starts from oracle prototypes, its advantage reflects the value of accurate prototype initialisation. These results suggest that prototype estimation and boundary estimation address complementary aspects of the estimation problem: a reliable prototype provides a strong starting point, while subsequent classifier refinement can further exploit information contained in the query distribution.

\begin{figure}
\centering
\includegraphics[width=0.81\linewidth]{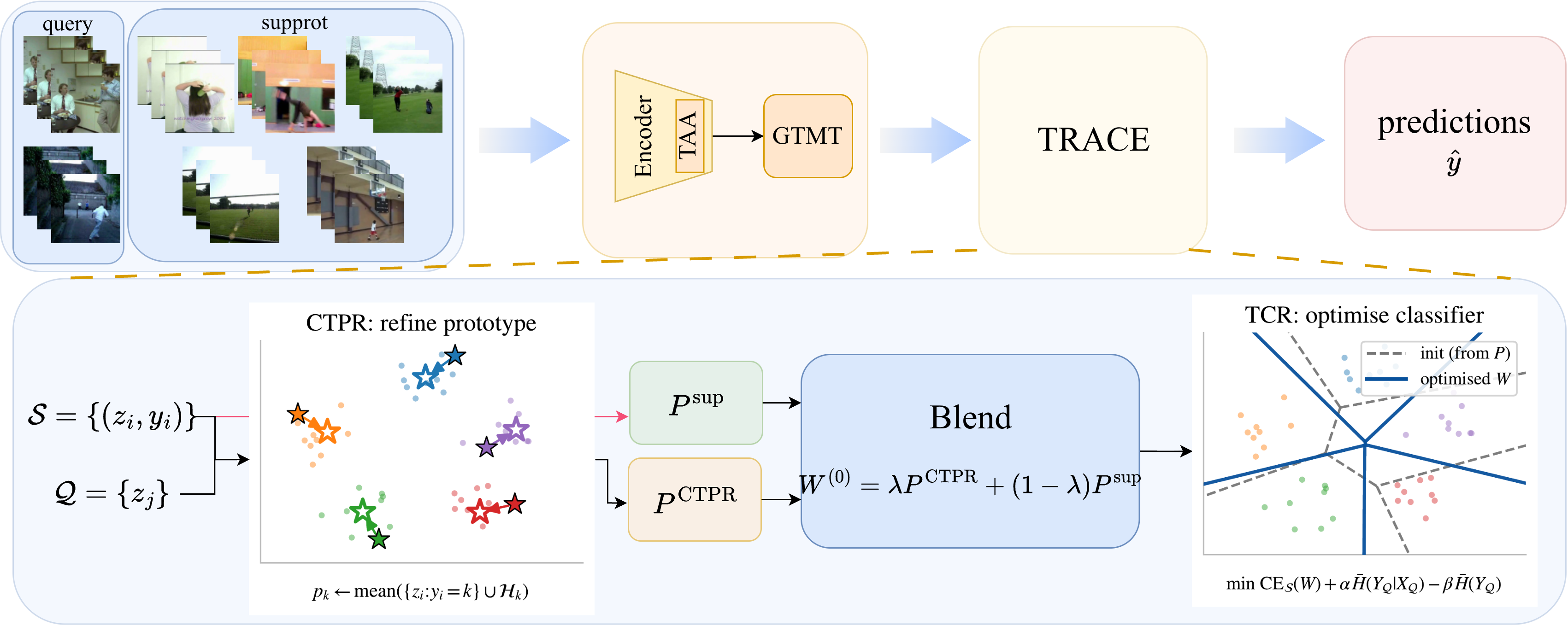}
\caption{\textbf{\sotd{}: recovering the estimation gap at test time.}
\textbf{Top:} a target-domain episode is encoded by a frozen VideoMAE-S encoder equipped with Temporal-Aware Adapters (TAA) and a Global Temporal-aware Moment Tuning (GTMT) head. \textbf{Bottom:} \sotd{} first applies Confidence-Thresholded Transductive Prototype Refinement (CTPR) to refine prototype estimates using only confident query samples, then performs Transductive Classifier Refinement (TCR) to adapt the decision boundary to the query distribution. The refined prototypes initialise the classifier before boundary refinement. Insets illustrate the effect of CTPR and TCR in a schematic feature space (stars: support; dots: query).}

\label{fig:pipeline}
\end{figure}

\section{\sotd{}: Test-Time Recovery of the Estimation Gap}
\label{sec:method}

The oracle analysis in Sec.~\ref{sec:analysis} showed that recovering the estimation gap requires addressing two complementary sources of error: prototype estimation and boundary estimation. It further shows that boundary refinement is most effective when initialised from an improved prototype estimate. These observations motivate \sotd{}, a training-free transductive inference procedure that operates entirely at test time while keeping the representation fixed.

Given the support and query videos, a frozen VideoMAE-S encoder equipped with Temporal-Aware Adapters (TAA) and a Global Temporal-aware Moment Tuning (GTMT) head first maps every clip into an embedding, after which all subsequent computation is performed within the episode while keeping the encoder fixed.

As illustrated in Fig.~\ref{fig:pipeline}, \sotd{} addresses the resulting estimation problem through two complementary stages. Confidence-Thresholded Transductive Prototype Refinement (CTPR) reduces prototype estimation error by incorporating only confidently predicted query samples into the prototype estimate. Transductive Classifier Refinement (TCR) then adapts a lightweight per-episode classifier to the structure of the query distribution, improving boundary estimation. Rather than treating prototype refinement and boundary refinement as independent improvements, \sotd{} uses the refined prototypes produced by CTPR to initialise TCR. This refined-prototype initialisation directly follows the insight provided by the Initialisation Oracle: boundary refinement is most effective when it starts from an improved prototype estimate rather than from the raw support prototypes alone.

\subsection{Confidence-Thresholded Transductive Prototype Refinement (CTPR)}
\label{sec:ctpr}

The oracle analysis in Sec.~\ref{sec:analysis} showed that a substantial portion of the estimation gap arises from inaccurate prototype estimation. The Prototype Oracle indicates that more accurate prototype estimates can substantially improve recognition accuracy. CTPR seeks to approximate this oracle behaviour at test time by refining support prototypes using confidently pseudo-labelled query samples.

Under severe domain shift, incorrect pseudo-labels can easily move a prototype away from its class centre. CTPR therefore prioritises precision over coverage: only highly confident queries are incorporated into the prototype estimate. Starting from the support prototypes of Eq.~\eqref{eq:proto}, we iteratively score all query samples, assign pseudo-labels, and retain only those whose confidence exceeds a threshold $\tau$.

\begin{equation}
\label{eq:ctpr-score}
\hat y_j^{(t)} = \arg\max_k \mathrm{sim}\!\big(z_j, p_k^{(t-1)}\big),
\qquad
c_j^{(t)} = \max_k \;\sm\!\big(\mathrm{sim}(z_j, p^{(t-1)})\big)_k ,
\end{equation}
This gives the high-confidence set $\mathcal{H}_k^{(t)}=\{\,j : \hat
y_j^{(t)}=k,\; c_j^{(t)}>\tau\,\}$. Each prototype is then recomputed as the mean of its support embeddings and selected query embeddings,
\begin{equation}
\label{eq:ctpr-refine}
p_k^{(t)} \;=\; \mean\!\Big( \{\,z_i : y_i=k\,\}\;\cup\;\{\,z_j : j\in\mathcal{H}_k^{(t)}\,\} \Big).
\end{equation}

After $T$ refinement rounds, we obtain the refined prototypes $p_k^{\textsc{ctpr}}$. CTPR deliberately adopts a hard confidence threshold and simple averaging. This keeps the procedure parameter-free, avoids fitting additional parameters from extremely limited support data, and makes refinement depend only on highly reliable query evidence.

\subsection{Transductive Classifier Refinement (TCR)}
\label{sec:tcr}

While CTPR improves prototype estimation, it still classifies queries through nearest-prototype matching. The Classifier Oracle in Sec.~\ref{sec:analysis} showed that additional gains remain available through improved boundary estimation. TCR addresses this by adapting a lightweight classifier to the structure of the query distribution.

Unlike prototype refinement, boundary refinement requires exploiting the global structure of the query set rather than only re-estimating class centres. Following the transductive information-maximisation framework of Boudiaf~\etal~\cite{boudiaf2020}, TCR optimises a per-episode linear classifier initialised from a prototype matrix.
\begin{equation}
\label{eq:tcr-logit}
\ell_k(z;W) = \gamma\,\frac{\langle z, W_k\rangle}{\lVert z\rVert\,\lVert W_k\rVert},
\qquad
P(k\mid z) = \sm\!\big(\ell(z;W)\big)_k .
\end{equation}
The objective fits the labelled supports while encouraging the query predictions to be confident and class-balanced:
\begin{equation}
\label{eq:tcr-loss}
\mathcal{L}(W) \;=\; \mathrm{CE}_{\Sset}(W)
\;+\;\alpha\,\bH(Y_\Qset\!\mid\! X_\Qset)
\;-\;\beta\,\bH(Y_\Qset),
\end{equation}
where $\mathrm{CE}_{\Sset}(W)=-\tfrac{1}{|\Sset|}\sum_{(x_i,y_i)\in\Sset}\log P(y_i\mid z_i)$
is the support cross-entropy, and
\begin{align}
\label{eq:tcr-ent}
\bH(Y_\Qset\!\mid\! X_\Qset) &= -\frac{1}{|\Qset|}\sum_{j}\sum_k P(k\mid z_j)\log P(k\mid z_j), \\
\bH(Y_\Qset) &= -\sum_k \bar p_k\log \bar p_k,\quad \bar p_k=\frac{1}{|\Qset|}\sum_j P(k\mid z_j).
\end{align}

The conditional-entropy term encourages confident predictions, while the marginal-entropy term discourages degenerate solutions in which most queries collapse to a small subset of classes. Together, the two terms encourage confident predictions while maintaining class balance across the query set. Since $\bH(Y_\Qset)-\bH(Y_\Qset\!\mid\! X_\Qset)$ equals the empirical mutual information between queries and predicted labels, Eq.~\eqref{eq:tcr-loss} can be viewed as maximising query mutual information while fitting the support set. Only the classifier parameters are updated within the episode; the feature representation remains unchanged throughout.

\begin{algorithm}[t]
\caption{\sotd{}: training-free recovery of the estimation gap within a test episode.}
\label{alg:trace}
\begin{algorithmic}[1]
\Require support embeddings ${(z_i,y_i)}$, query embeddings ${z_j}$;
hyperparameters $\tau,T,\lambda,\alpha,\beta,\gamma$

\State $P_k^{\mathrm{sup}} \gets \frac{1}{N}\sum_{i:y_i=k} z_i$;;
$p_k \gets P_k^{\mathrm{sup}}$
\Comment{support prototypes}

\For{$t=1$ \textbf{to} $T$} \Comment{CTPR}
\State Assign pseudo-labels and confidence scores to all queries using ${p_k}_{k=1}^{K}$ (Eq.~\ref{eq:ctpr-score})
\State Collect $\mathcal{H}_k^{(t)}$, queries pseudo-labelled $k$ with confidence $>\tau$
\If{$\bigcup_k \mathcal{H}_k^{(t)}=\emptyset$}
\textbf{continue}
\EndIf
\State $p_k \gets \mean({z_i:y_i=k}\cup{z_j:j\in\mathcal{H}_k^{(t)}})$
\Comment{Eq.~\ref{eq:ctpr-refine}}
\EndFor

\State $P^{\textsc{ctpr}}\gets[p_1,\dots,p_K]$
\State $W \gets \lambda P^{\textsc{ctpr}} + (1-\lambda)P^{\mathrm{sup}}$
\Comment{refined-prototype initialisation, Eq.~\ref{eq:trace-init}}

\For{$100$ steps} \Comment{TCR}
\State $W \gets \mathrm{AdamStep}(W,\nabla_W\mathcal{L}(W))$
\Comment{Eq.~\ref{eq:tcr-loss}}
\EndFor

\State \Return $\arg\max_k \ell_k(z_j;W)$ for each query $j$
\end{algorithmic}
\end{algorithm}

\subsection{\sotd{}: Refined-Prototype Initialisation}
\label{sec:trace}

The oracle analysis further showed that prototype refinement and boundary refinement should not be treated as independent improvements. The Initialisation Oracle consistently provides the strongest or near-strongest oracle performance, indicating that classifier refinement benefits most from starting from a high-quality prototype estimate.

Motivated by this observation, \sotd{} uses the CTPR-refined prototypes to initialise TCR. Rather than optimising the classifier from raw support prototypes alone, we initialise it from a convex combination of the raw and refined prototypes:

\begin{equation}
\label{eq:trace-init}
W^{(0)} \;=\; \lambda\,P^{\textsc{ctpr}} \;+\; (1-\lambda)\,P^{\mathrm{sup}},
\end{equation}
where $P^{\mathrm{sup}},P^{\textsc{ctpr}}\!\in\!\R^{K\times d}$ stack the raw and CTPR-refined prototypes, with $k$-th rows $p_k$ and $p_k^{\textsc{ctpr}}$, respectively.  TCR is then optimised from this initialisation using Eq.~\eqref{eq:tcr-loss}. The resulting design allows boundary refinement to inherit the robustness of the refined prototypes while retaining the flexibility of classifier adaptation. In this sense, \sotd{} is not merely a sequential combination of CTPR and TCR, but a realisation of the refined-prototype initialisation strategy identified by the Initialisation Oracle. Algorithm~\ref{alg:trace} summarises the complete procedure.

\begin{table}
\centering\small
\renewcommand{\arraystretch}{0.92}
\begin{tabular}{lcccccc}
\toprule
Method & HMDB51 & UCF101 & SSv2 & Diving48 & RareAct & Avg. \\
\midrule
\multicolumn{7}{l}{\emph{Source: Kinetics-400 \quad(5-way 1-shot)}}\\
TAMT~\cite{tamt2025}$^{\ddagger}$ & 58.19 & 86.04 & 41.14 & 32.36 & 46.33 & 52.81 \\
\sotd{} (ours)                    & \textbf{64.81} & \textbf{95.35} & \textbf{43.56} & \textbf{34.15} & \textbf{51.09} & \textbf{57.79} \\
\midrule
\multicolumn{7}{l}{\emph{Source: Kinetics-400 \quad(5-way 5-shot)}}\\
STARTUP++~\cite{startup2021} & 44.71 & 60.82 & 39.60 & 14.92 & 45.22 & 41.05 \\
DD++~\cite{dd2021}           & 48.04 & 63.26 & 44.50 & 16.23 & 47.01 & 43.81 \\
STRM~\cite{strm2022}         & 24.98 & 42.33 & 35.01 & 16.69 & 39.01 & 31.60 \\
HyRSM~\cite{hyrsm2022}       & 29.81 & 45.65 & 40.09 & 17.57 & 44.27 & 35.49 \\
CDFSL-V~\cite{cdfslv2023}    & 53.23 & 65.42 & 49.92 & 17.84 & 49.80 & 47.24 \\
TAMT~\cite{tamt2025}$^{\ddagger}$ & 72.50 & 95.96 & 59.28 & 44.91 & 68.18 & 68.17 \\
\sotd{} (ours)               & \textbf{74.23} & \textbf{97.43} & \textbf{61.45} & \textbf{45.55} & \textbf{69.67} & \textbf{69.67} \\
\midrule
\multicolumn{7}{l}{\emph{Source: Kinetics-100 \quad(5-way 1-shot)}}\\
STARTUP++~\cite{startup2021} & 16.66 & 24.48 & 14.17 & 13.13 & 17.21 & 17.13 \\
DD++~\cite{dd2021}           & 17.44 & 26.04 & 14.96 & 13.73 & 19.02 & 18.24 \\
CDFSL-V~\cite{cdfslv2023}    & 18.59 & 27.78 & 16.01 & 14.11 & 20.06 & 19.31 \\
SEEN$^{*\dagger}$~\cite{seen2023} & 39.50 & 64.60 & 26.20 & 31.23 & 34.28 & 39.16 \\
DMSD$^{*\dagger}$~\cite{dmsd2024} & 39.62 & 65.00 & 26.78 & 34.20 & 37.50 & 40.62 \\
TAMT~\cite{tamt2025}$^{\ddagger}$         & 47.02 & 72.38 & 34.45 & 27.04 & 36.04 & 43.39 \\
\sotd{} (ours)               & \textbf{52.81} & \textbf{80.27} & \textbf{36.62} & \textbf{28.76} & \textbf{40.89} & \textbf{47.87} \\
\midrule
\multicolumn{7}{l}{\emph{Source: Kinetics-100 \quad(5-way 5-shot)}}\\
STARTUP++~\cite{startup2021} & 24.97 & 32.20 & 15.16 & 14.55 & 31.77 & 23.73 \\
DD++~\cite{dd2021}           & 25.99 & 34.10 & 16.00 & 16.24 & 31.20 & 24.71 \\
CDFSL-V~\cite{cdfslv2023}    & 29.80 & 36.53 & 17.21 & 16.37 & 33.91 & 26.76 \\
SEEN$^{*\dagger}$~\cite{seen2023} & 52.80 & 79.60 & 31.20 & 40.90 & 50.20 & 50.94 \\
DMSD$^{*\dagger}$~\cite{dmsd2024} & 54.90 & 81.90 & 32.10 & 42.28 & 53.30 & 52.90 \\
TAMT~\cite{tamt2025}$^{\ddagger}$         & 61.76 & 87.76 & 48.90 & 38.33 & 52.81 & 57.91 \\
\sotd{} (ours)               & \textbf{62.99} & \textbf{89.11} & \textbf{51.23} & \textbf{38.83} & \textbf{54.08} & \textbf{59.25} \\
\bottomrule
\end{tabular}
\caption{\textbf{Main results across five CD-FSAR target datasets} (5-way accuracy, 
Methods marked with $^{*}$ or $^{\dagger}$ use different input resolutions or backbones and are included for reference. All other methods follow the common VideoMAE ViT-S/$112$ protocol. $^{\ddagger}$ denotes our reproduction of TAMT, the base model on which \sotd{} is applied. Prior work does not report K-400 1-shot results; therefore only TAMT and \sotd{} are shown in this setting. Best results under the common ViT-S/$112$ protocol are highlighted in \textbf{bold}.}

\label{tab:main}
\end{table}

\section{Experiments}
\label{sec:exp}

\subsection{Setup}

\textbf{Datasets and source domains.}
We evaluate on five standard CD-FSAR target datasets: HMDB51, UCF101, SSv2, Diving48, and RareAct. Following prior work, we report results under two source domains: \emph{Kinetics-400} (with target-overlapping classes removed, leaving 364 categories~\cite{tamt2025}) and \emph{Kinetics-100}, the smaller source domain used by SEEN~\cite{seen2023} and DMSD~\cite{dmsd2024}. We adopt the CDFSL-V splits~\cite{cdfslv2023} for HMDB51, UCF101, SSv2, and Diving48, and the TAMT 48/8/8 base/validation/novel split for RareAct~\cite{tamt2025}. Source and target classes are disjoint in all settings.

\noindent\textbf{Protocol and evaluation.}
Following~\cite{tamt2025,cdfslv2023}, we evaluate 5-way 1-shot and 5-shot classification with 12 query videos per class. Each video is represented by 16 sampled frames at $112\times112$ resolution. We report mean accuracy over 10,000 test episodes (5 splits $\times$ 2,000 episodes), matching the evaluation protocol of TAMT~\cite{tamt2025}. The baseline predicts each query by matching it to support prototypes using inner-product similarity in the 1-shot setting and negative squared Euclidean distance in the 5-shot setting. 

\noindent\textbf{Backbone.}
The main experiments build on TAMT~\cite{tamt2025}, a strong CD-FSAR baseline consisting of a frozen VideoMAE-S~\cite{videomae2022} backbone augmented with Temporal-Aware Adapters (TAA) and a Global Temporal-aware Moment Tuning (GTMT) head. Since \sotd{} operates entirely at test time, it does not modify the backbone or episodic head. Using a strong baseline reduces the risk that improvements arise from weaknesses in the representation.

\noindent\textbf{\sotd{} configuration and implementation.}
Unless stated otherwise, all experiments use a single configuration across datasets and shot settings: CTPR with confidence threshold $\tau=0.5$ and $T=3$ refinement rounds; TCR with $\alpha=0.1$, $\beta=1.0$, temperature $\gamma=15$, and 100 Adam optimisation steps at learning rate $10^{-3}$; and refined-prototype blending weight $\lambda=0.5$. No per-dataset hyperparameter tuning is performed. Sec.~\ref{sec:exp-ablation} analyses the sensitivity of these choices and shows that a single configuration performs consistently across datasets.

We implement \sotd{} in PyTorch and evaluate using the publicly released TAMT checkpoints~\cite{tamt2025} for both Kinetics-400 and Kinetics-100 source domains. Experiments are conducted on a single NVIDIA H200 GPU. Relative to standard prototype inference, \sotd{} adds approximately 100,ms per episode, with the TCR optimisation accounting for most of the overhead.


\subsection{Comparison with the State of the Art}

Table~\ref{tab:main} reports the main CD-FSAR results across five target datasets, two source domains, and both 1-shot and 5-shot settings. Across all settings, \sotd{} consistently improves its TAMT baseline despite requiring no additional training and modifying neither the backbone nor the episodic head.

Under the Kinetics-400 source domain, \sotd{} improves every target dataset. The gains are particularly pronounced in the 1-shot regime, where average accuracy increases by roughly five points and exceeds nine points on UCF101. Similar improvements are observed under the Kinetics-100 source domain, where \sotd{} consistently outperforms TAMT across all datasets and shot settings. Under the common ViT-S/$112$ protocol, \sotd{} achieves the best result in every Kinetics-100 setting.

A clear pattern emerges across both source domains: improvements are substantially larger in the 1-shot setting than in the 5-shot setting. This trend is consistent with the oracle analysis in Sec.~\ref{sec:analysis}, which showed that the estimation gap grows as support becomes scarcer. The largest gains therefore occur precisely in the settings where estimation errors are expected to be most severe. Taken together, these results indicate that a substantial portion of the remaining performance gap can be recovered through improved episode-level estimation at test time, without modifying the representation or training procedure.

\begin{table}
\centering\small
\begin{tabular}{llcccc}
\toprule
Dataset & Shot & Baseline & +CTPR & +TCR & +\sotd{} \\
\midrule
HMDB51   & 1 & 58.19 & 61.82 & 63.50 & \textbf{64.81} \\
         & 5 & 72.50 & 73.95 & 73.25 & \textbf{74.23} \\
UCF101   & 1 & 86.04 & 92.73 & 93.44 & \textbf{95.35} \\
         & 5 & 95.96 & 96.48 & \textbf{97.47} & 97.43 \\
SSv2     & 1 & 41.14 & 42.74 & 43.12 & \textbf{43.56} \\
         & 5 & 59.28 & 60.76 & 60.50 & \textbf{61.45} \\
Diving48 & 1 & 32.36 & 32.74 & 33.96 & \textbf{34.15} \\
         & 5 & 44.91 & 45.35 & 45.17 & \textbf{45.55} \\
RareAct  & 1 & 46.33 & 49.12 & 50.02 & \textbf{51.09} \\
         & 5 & 68.18 & 68.90 & \textbf{70.26} & 69.67 \\
\bottomrule
\end{tabular}
\caption{\textbf{Ablation of prototype and boundary refinement} (5-way accuracy, 
Baseline denotes the original TAMT inference procedure. \emph{$+$CTPR} applies prototype refinement only, \emph{$+$TCR} applies classifier refinement only, and \emph{$+\sotd{}$} combines the two through refined-prototype initialisation. Best results in each row are shown in \textbf{bold}.}

\label{tab:ablation}
\end{table}

\subsection{Ablation of Prototype and Boundary Refinement}
\label{sec:exp-ablation}

Table~\ref{tab:ablation} evaluates the contribution of the two refinement stages and their composition. Both CTPR and TCR consistently improve over the TAMT baseline, indicating that prototype refinement and boundary refinement provide complementary benefits. Their combination typically achieves the strongest performance; in the remaining two settings, the difference between \sotd{} and the strongest individual component is less than 0.6 points.

The gains are particularly pronounced in the 1-shot regime, where \sotd{} improves upon the baseline by between 1.8 and 9.3 points across datasets. This trend is consistent with the oracle analysis in Sec.~\ref{sec:analysis}, which showed that the estimation gap becomes larger as support becomes scarcer. It is also consistent with the finding that refined-prototype initialisation is the most effective way to combine prototype and boundary estimation.

\begin{table}
\centering\footnotesize
\input{tables/gentable_T4_1shot}
\caption{\textbf{Generalisation across episodic heads} (5-way 1-shot accuracy, 
\sotd{} is applied unchanged to a ProtoNet head and a good-embedding linear head. Best results within each head are highlighted in \textbf{bold}.}

\label{tab:gen}
\end{table}

\subsection{Generality Across Episodic Heads}
\label{sec:exp-generalisation}

To assess whether the gains of \sotd{} depend on a particular episodic head, we apply it unchanged to ProtoNet~\cite{protonet2017} and a good-embedding linear head~\cite{goodembed2020}, in addition to the TAMT head used throughout the main experiments.

Table~\ref{tab:gen} shows that \sotd{} consistently improves all three heads. Across the five target datasets, it increases the average 1-shot accuracy by 4.5 points on ProtoNet and 3.5 points on the good-embedding head, while achieving the strongest average performance in both cases. The gains are observed across nearly all datasets, indicating that the benefit of \sotd{} is not tied to a particular head design.

The effectiveness of \sotd{} across substantially different heads is consistent with the central hypothesis of this work. Because \sotd{} operates entirely through test-time estimation using the support and query sets, its gains do not rely on head-specific architectural choices. The results therefore suggest that the recovered performance stems from improved episode-level estimation rather than from a particular classifier design. 

\begin{figure}
\centering
\includegraphics[width=0.59\linewidth]{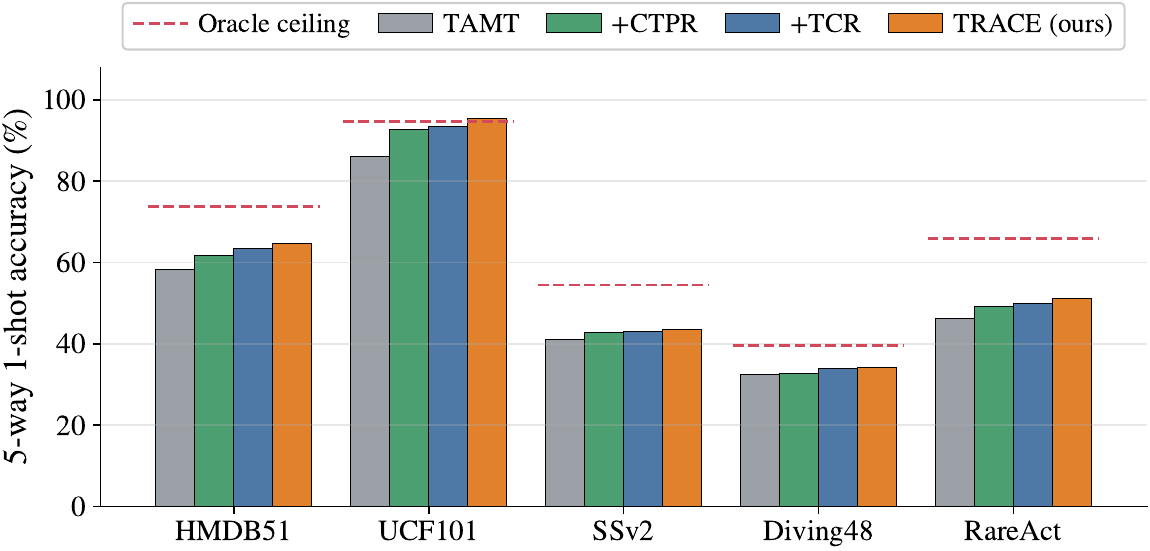}
\includegraphics[width=0.355\linewidth]{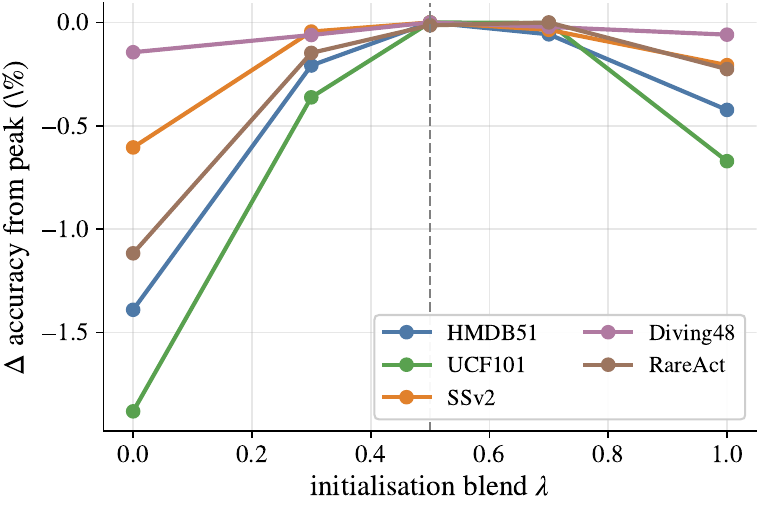}
\caption{\textbf{Recovering the estimation gap through refined-prototype initialisation.}
\textbf{Left:} comparison of TAMT, CTPR, TCR, and \sotd{} against the corresponding oracle ceiling for each dataset (dashed lines; Table~\ref{tab:oracle}).
\textbf{Right:} effect of the initialisation blend $\lambda$. $\lambda=0$ corresponds to TCR initialised from the raw support prototype, while $\lambda=1$ corresponds to using only the refined prototype from CTPR. Curves show the accuracy difference from each dataset's peak performance, allowing all datasets to be compared on a common scale.
}
\label{fig:gap_closure}
\end{figure}

\subsection{Mechanism and Robustness Analysis}

Figure~\ref{fig:gap_closure} (left) compares the gains of \sotd{} with the oracle gaps identified in Sec.~\ref{sec:analysis}. Without any additional training, \sotd{} recovers a substantial fraction of the available headroom on every target dataset, improving upon the baseline by 1.8--9.3 points. The remaining headroom is largest on SSv2 and Diving48, suggesting greater scope for future improvements on highly temporal actions. Figure~\ref{fig:gap_closure} (right) examines the refined-prototype initialisation that links CTPR and TCR. Performance peaks consistently around $\lambda=0.5$, while performance decreases toward either extreme. In particular, $\lambda=0$ reduces to TCR initialised from the raw support prototypes, whereas $\lambda=1$ relies entirely on the refined prototypes. This behaviour is consistent with the Initialisation Oracle in Sec.~\ref{sec:analysis}, which identified refined-prototype initialisation as the most effective way to combine prototype refinement and boundary refinement.

\begin{figure}
\centering
\includegraphics[width=0.86\linewidth]{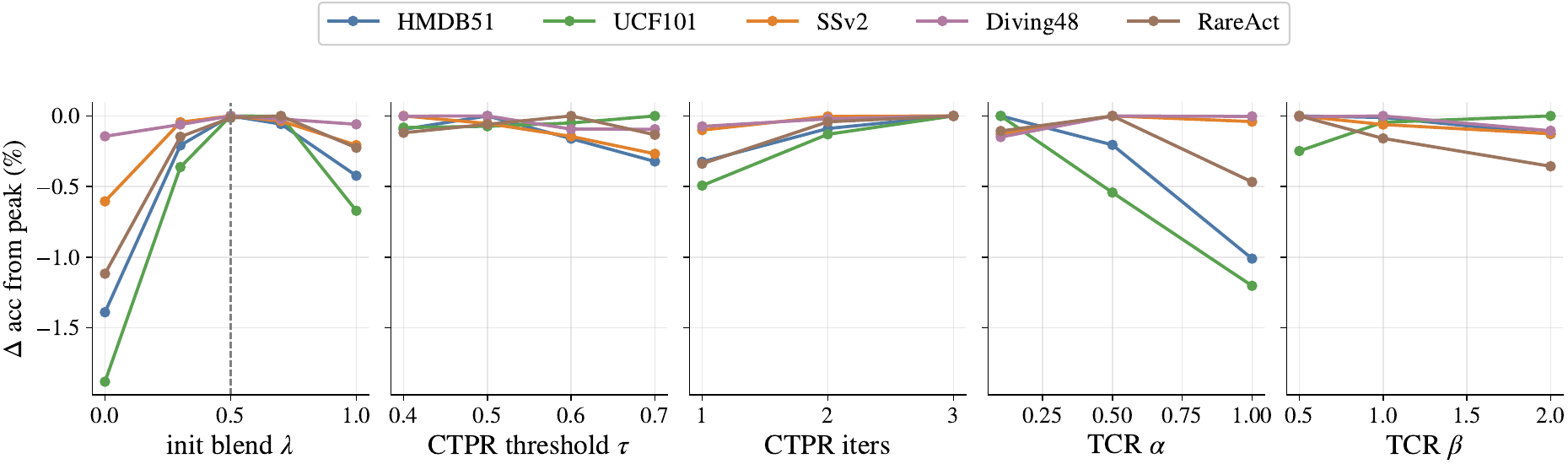}
\includegraphics[width=0.86\linewidth]{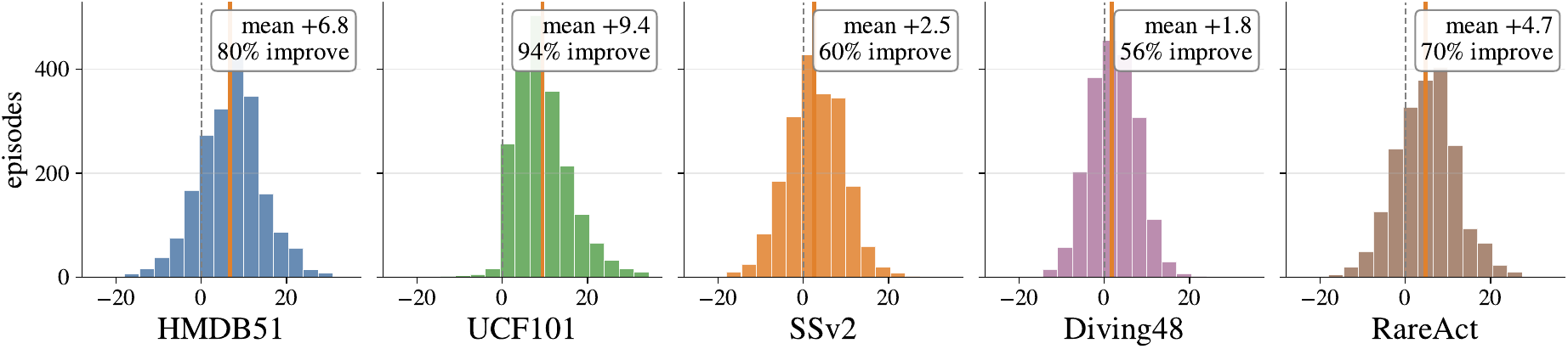}
\caption{\textbf{Robustness to hyperparameters and consistency across episodes.}
\textbf{Top:} sensitivity to $\tau$, $T$, $\alpha$, and $\beta$ around the default configuration. Curves show the accuracy difference from each dataset's peak performance, allowing all datasets to be compared on a common scale.
\textbf{Bottom:} per-episode distribution of \sotd{} accuracy minus baseline accuracy over $1{,}000$ episodes per dataset. Each panel reports the mean improvement and the proportion of episodes that outperform the baseline.
}
\label{fig:hparam}
\end{figure}

Figure~\ref{fig:hparam} (top) further shows broad performance plateaus around the default configuration, indicating that \sotd{} is not sensitive to precise hyperparameter choices. Across all datasets, the default setting lies close to the optimum and no dataset-specific tuning is required. Figure~\ref{fig:hparam} (bottom) plots the distribution of per-episode accuracy changes relative to the baseline. The majority of episodes improve, while only a small minority experience a decrease, indicating that the average improvement reflects a broad shift across episodes rather than a few favourable cases. 

\section{Discussion}
\label{sec:discussion}

\paragraph{Shot dependence of the estimation gap.}
\sotd{} delivers its largest gains in the low-shot regime, which is exactly
what the estimation-gap hypothesis predicts. Repeating the oracle analysis of
Sec.~\ref{sec:analysis} at 5-shot shows that the gap itself contracts
substantially relative to 1-shot. With more labelled examples, prototypes
and decision boundaries are already estimated more accurately from the
support set, leaving less room for test-time refinement. The reduced benefit
at 5-shot is therefore not a limitation of \sotd{}, but a direct consequence
of the bottleneck becoming less severe. More generally, the recoverable gain
depends on two factors: the size of the estimation gap and the reliability of
the pseudo-labels used for refinement. When either becomes small, little
headroom remains for test-time adaptation.

\begin{figure}
\centering
\includegraphics[width=0.79\linewidth]{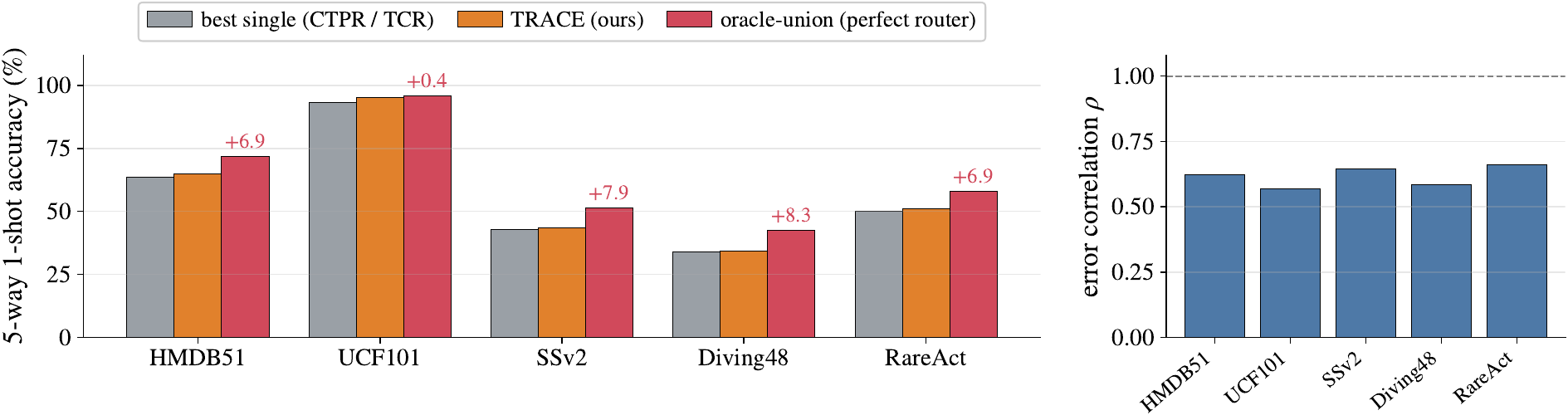}
\caption{\textbf{Initialisation succeeds where routing fails.}
\textbf{Left:} per dataset, the left bar shows the better of CTPR and TCR, the centre bar shows \sotd{}, and the right bar shows the accuracy of a perfect per-episode router (oracle routing), representing the maximum benefit obtainable from selecting between the two estimators. \textbf{Right:} per-episode error correlation $\rho$ between CTPR and TCR. Higher correlation indicates that the two estimators tend to fail on the same episodes.
}
\label{fig:comp}
\end{figure}

\paragraph{Initialisation versus routing.}
CTPR and TCR are complementary, but their complementarity is difficult to
exploit directly. Figure~\ref{fig:comp} (left) shows that a perfect
per-episode router could substantially outperform \sotd{}, indicating that
additional complementarity remains available in principle. However,
Figure~\ref{fig:comp} (right) shows that the errors of CTPR and TCR are
strongly correlated ($\rho\!\approx\!0.6$), and we found no reliable
label-free routing signal that predicts which estimator will perform better
on a given episode. As a result, confidence-based routing recovers only a
small fraction of the available headroom.

Taken together, the two panels suggest that the limitation is not a lack of
complementarity, but the absence of a reliable routing signal. Rather than
selecting between two competing estimators, \sotd{} combines them through
refined-prototype initialisation, allowing information from both estimators
to influence the final decision boundary without requiring an explicit
routing decision. This makes the combination substantially more robust than
the routing schemes we explored. Closing the remaining oracle-routing gap
therefore appears to require new forms of label-free episode-level
uncertainty estimation, which we regard as an interesting direction for
future work.

\paragraph{Limits of additional complexity.}
We also explored several alternatives aimed at exploiting the residual
complementarity between CTPR and TCR. None produced consistent improvements
over \sotd{}. This negative
result is consistent with the analysis above. Although some complementary
information remains, it is not readily separable using the
unlabelled signal accessible at test time. Additional machinery therefore
adds complexity without reliably recovering further estimation gains. We
consequently retain the simplest design that consistently realises the
estimation-gap reductions identified by the oracle analysis.

\section{Conclusion}
\label{sec:conclusion}

We revisited cross-domain few-shot action recognition through the lens of the estimation gap. Our oracle analysis showed that, even with strong frozen representations, a substantial portion of the remaining error arises from estimating prototypes and decision boundaries from extremely limited support data. Crucially, the unlabelled query set contains sufficient signal to recover much of this gap at test time.
Based on this observation, we introduced \sotd{}, a training-free transductive inference procedure that improves episode-level estimation using only the data available at test time. Across datasets, shot settings, and episodic heads, \sotd{} consistently recovers a substantial fraction of the gap predicted by the oracle analysis.
More broadly, representation quality is not the only bottleneck in CD-FSAR: under severe support scarcity, better episode-level estimation is an effective and portable complement to representation learning. We hope the estimation-gap perspective proves useful for other cross-domain few-shot problems.

\section*{Acknowledgements}
This work was funded by Business Finland (grant No 69/31/2025) under the project Human-centric AI-enabled Collaborative Fieldwork Operations (AiWo). The calculations were performed using computational resources provided by the Aalto University Science-IT project.

\bibliography{egbib}
\end{document}


\maketitle

The main paper focuses on the 5-way 1-shot regime, where the estimation gap is largest and the effect of test-time estimation is most pronounced. This supplementary document reports the corresponding 5-shot analyses to verify that the conclusions of the main paper remain consistent as additional labelled support examples become available. Specifically, we revisit the oracle analysis, gap-closure analysis, and robustness studies under the 5-shot setting. All analyses support the same conclusions as the 1-shot results reported in the main paper. In addition, Sec.~\ref{sec:supp-robustness} compares \sotd{} with established transductive inference methods on identical frozen features and evaluates robustness to class-imbalanced query sets and to the query budget.

\section{The Estimation Gap Contracts at 5-Shot}

The oracle analysis confirms the central observation of the main paper: the estimation gap remains present at 5-shot, but is substantially smaller than in the 1-shot regime. As additional support examples become available, prototype and boundary estimates become more reliable, reducing the headroom available for test-time refinement.

\begin{figure}[h]
\centering
\includegraphics[width=0.8\linewidth]{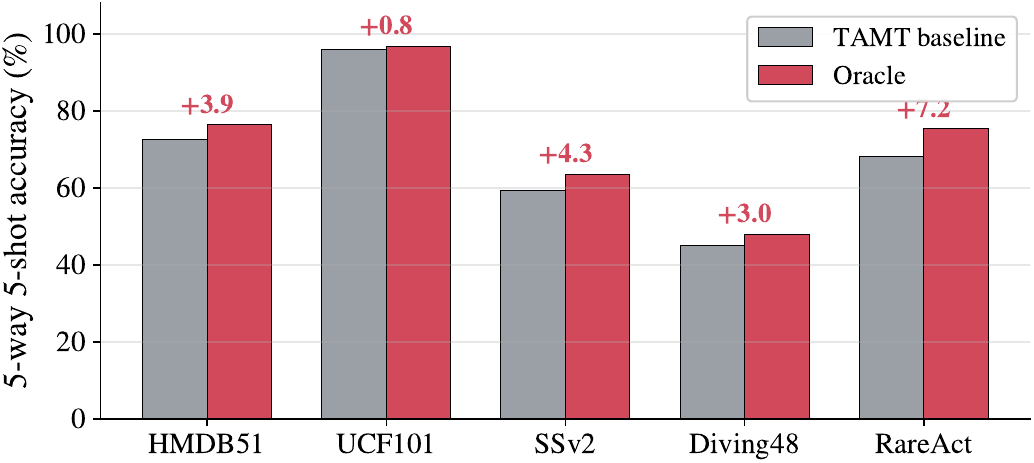}
\caption{\textbf{Baseline and oracle performance in the 5-way 5-shot setting.}
Numbers above the bars indicate the oracle gap. Oracle performance corresponds to the strongest of the three transductive upper bounds in Table~\ref{tab:oracle5}.}
\label{fig:gap5}
\end{figure}

\begin{figure}[h]
\centering
\includegraphics[width=0.8\linewidth]{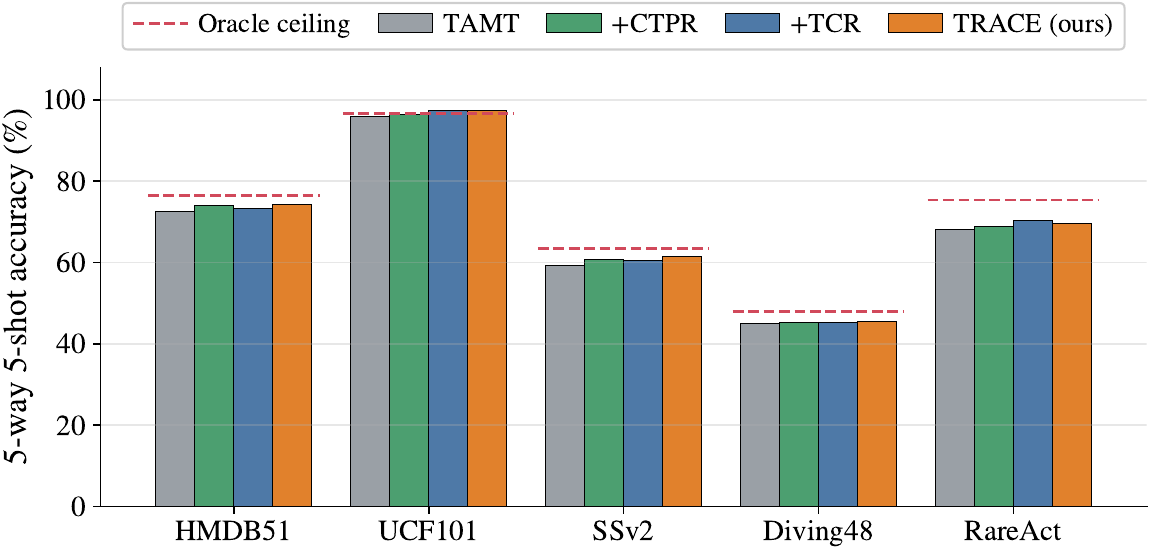}
\caption{\textbf{Recovering the reduced estimation gap at 5-shot.}
Progressive improvement from the TAMT baseline to +CTPR, +TCR, and \sotd{}, compared with the corresponding oracle ceiling for each dataset. Dashed lines indicate the oracle ceiling from Table~\ref{tab:oracle5}.}
\label{fig:gapclose5}
\end{figure}

Table~\ref{tab:oracle5} quantifies this effect. The oracle gap contracts to $0.8$--$7.2$ points, compared with $7.3$--$19.5$ points at 1-shot. As prototype estimates become increasingly reliable, TUB-P becomes the strongest oracle on four of the five datasets, while the advantage of refined-prototype initialisation (TUB-I) is substantially reduced. Figure~\ref{fig:gap5} visualises the smaller estimation gap, and Figure~\ref{fig:gapclose5} shows that \sotd{} continues to recover a meaningful fraction of the remaining headroom.

\begin{table}[h]
\centering\small
\begin{tabular}{lccccc}
\toprule
Dataset & Baseline & TUB-P & TUB-C & TUB-I & gap $\Delta$ \\
\midrule
HMDB51   & 72.49 & \textbf{76.42} & 73.47 & 76.00 & $+3.93$ \\
UCF101   & 95.98 & \textbf{96.77} & 96.65 & 95.09 & $+0.79$ \\
SSv2     & 59.19 & \textbf{63.50} & 61.38 & 62.53 & $+4.31$ \\
Diving48 & 44.91 & \textbf{47.93} & 40.62 & 46.65 & $+3.02$ \\
RareAct  & 68.16 & 73.68 & 75.14 & \textbf{75.39} & $+7.23$ \\
\bottomrule
\end{tabular}
\caption{\textbf{Oracle performance under the 5-way 5-shot setting.}
5-shot counterpart of Table~1 in the main paper. Gap $\Delta$ denotes the difference between the best oracle and the baseline. Best oracle per row is shown in \textbf{bold}.}
\label{tab:oracle5}
\end{table}

\section{Generalisation Persists at 5-Shot}

The generalisation behaviour observed in the main paper remains unchanged at 5-shot: \sotd{} continues to improve performance across diverse episodic heads, although the margins become smaller as the estimation gap contracts.

\begin{figure}[h]
\centering
\includegraphics[width=0.8\linewidth]{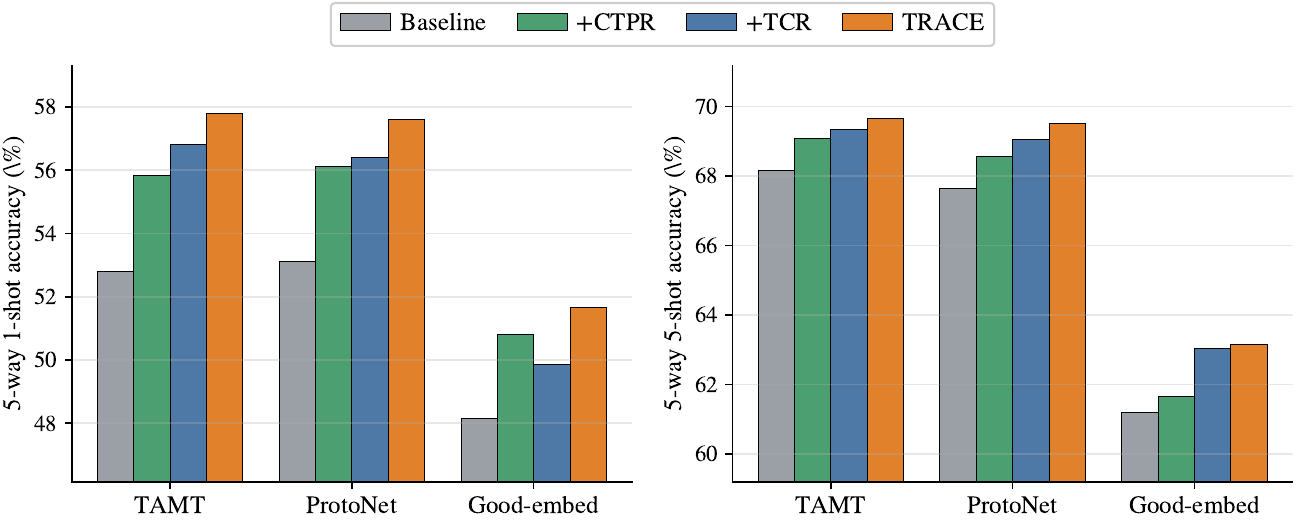}
\caption{\textbf{Average performance across episodic heads.}
Average accuracy over the five target datasets for the TAMT, ProtoNet, and Good-embed heads. Bars correspond to Baseline, +CTPR, +TCR, and \sotd{} under the 1-shot (left) and 5-shot (right) settings. The y-axis is truncated to highlight the relative differences between methods.}
\label{fig:gen}
\end{figure}

Table~\ref{tab:gen5} reports the per-dataset results, while Figure~\ref{fig:gen} summarises the average performance across datasets. Occasional reversals between TCR and \sotd{} on individual datasets are expected once the estimation gap becomes small. This behaviour is consistent with the oracle analysis: as the available headroom for test-time refinement decreases, the benefit of combining prototype and boundary refinement naturally becomes smaller, while remaining consistently positive on average. Across all evaluated heads, \sotd{} continues to achieve the strongest overall average performance.

\begin{table}[h]
\centering\footnotesize
\input{tables/gentable_T4_5shot}
\caption{\textbf{Generalisation across episodic heads at 5-shot.}
Results for ProtoNet and Good-embed under the 5-way 5-shot setting; the corresponding TAMT results are reported in the main-paper ablation study. Best method per column within each head is shown in \textbf{bold}.}
\label{tab:gen5}
\end{table}

\section{Robustness Persists at 5-Shot}

The robustness trends observed in the main paper remain unchanged at 5-shot. The default configuration continues to perform consistently across datasets despite the smaller estimation gap in this regime.

\begin{figure}[h]
\centering
\includegraphics[width=\linewidth]{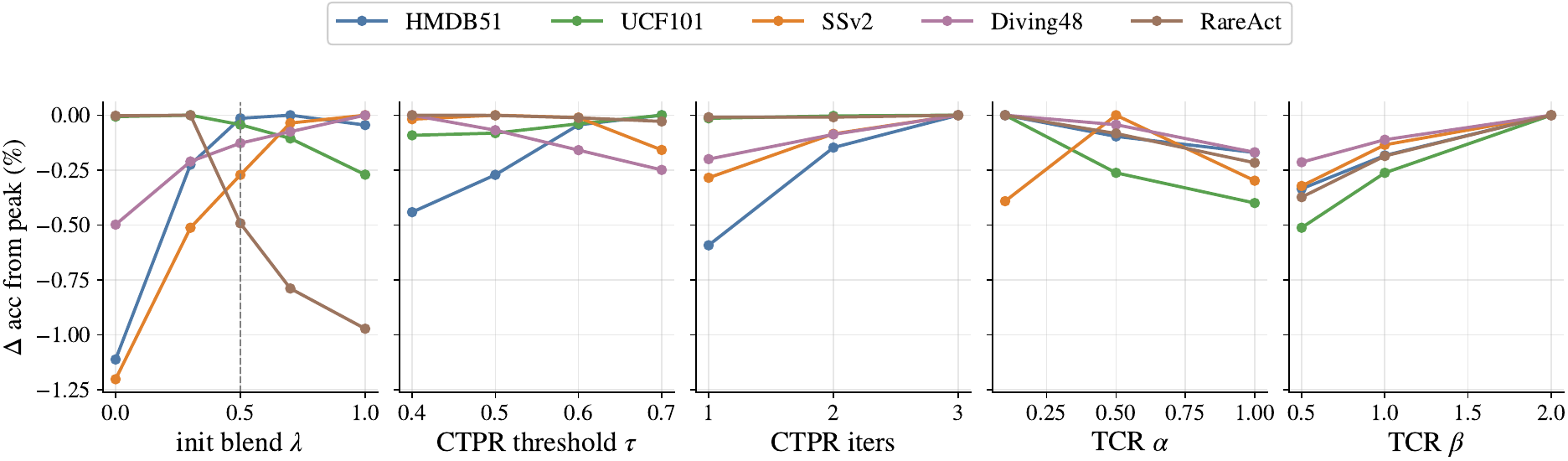}
\caption{\textbf{Sensitivity to hyperparameter choices at 5-shot.}
One-factor-at-a-time (OFAT) sweeps of $\tau$, $T$, $\alpha$, and $\beta$ around the default configuration. Curves show the accuracy difference from each dataset's peak performance, allowing all datasets to be compared on a common scale.
}
\label{fig:hparam5}
\end{figure}

\begin{figure}[h]
\centering
\includegraphics[width=0.55\linewidth]{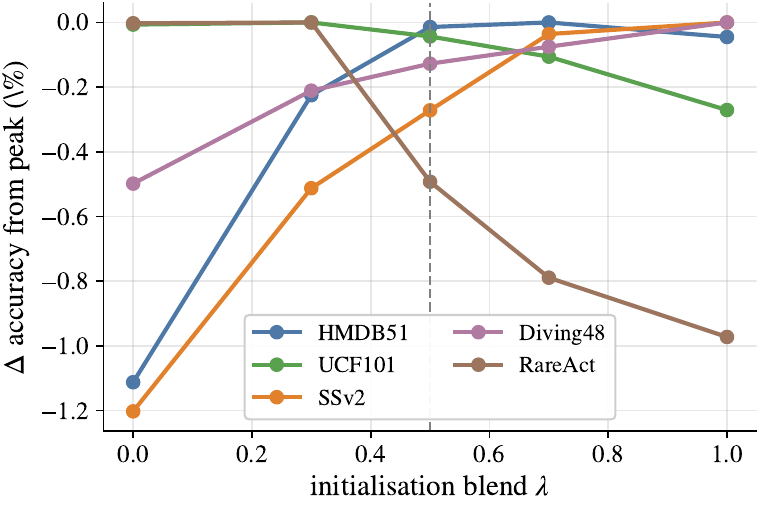}
\caption{\textbf{Sensitivity to the initialisation blend at 5-shot.}
Sweep of $\lambda$, where $\lambda=0$ corresponds to TCR initialised from the raw support prototype and $\lambda=1$ corresponds to using only the refined prototype from CTPR. Curves show the accuracy difference from each dataset's peak performance.
}
\label{fig:initw5}
\end{figure}

Figure~\ref{fig:hparam5} shows that the CTPR and TCR hyperparameters remain broadly insensitive around the default configuration. Figure~\ref{fig:initw5} revisits the refined-prototype initialisation strategy. The dependence on $\lambda$ becomes noticeably flatter than at 1-shot. This behaviour is consistent with the contraction of the estimation gap: as support prototypes become more reliable, the benefit of refined-prototype initialisation naturally decreases. It is also consistent with Table~\ref{tab:oracle5}, where the advantage of the refined-prototype oracle (TUB-I) is substantially reduced at 5-shot. Nevertheless, $\lambda=0.5$ remains a robust choice across datasets.

\section{Per-Episode Improvements and Complementarity Persist at 5-Shot}

The per-episode and complementarity analyses support the same conclusions as those reported in the main paper. Although the overall gains are smaller at 5-shot, improvements remain broadly distributed across episodes, and the complementary behaviour between CTPR and TCR persists.

\begin{figure}[h]
\centering
\includegraphics[width=\linewidth]{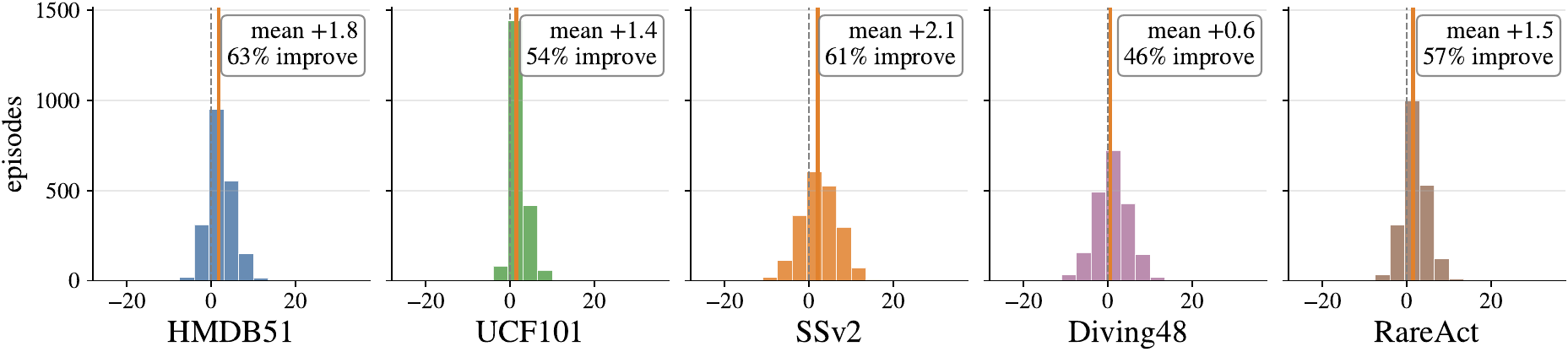}
\caption{\textbf{Per-episode improvements at 5-shot.}
Per-episode distribution of \sotd{} accuracy minus baseline accuracy over $1{,}000$ episodes per dataset. Each panel reports the mean improvement and the percentage of episodes that outperform the baseline.
}
\label{fig:gain5}
\end{figure}

\begin{figure}[h]
\centering
\includegraphics[width=\linewidth]{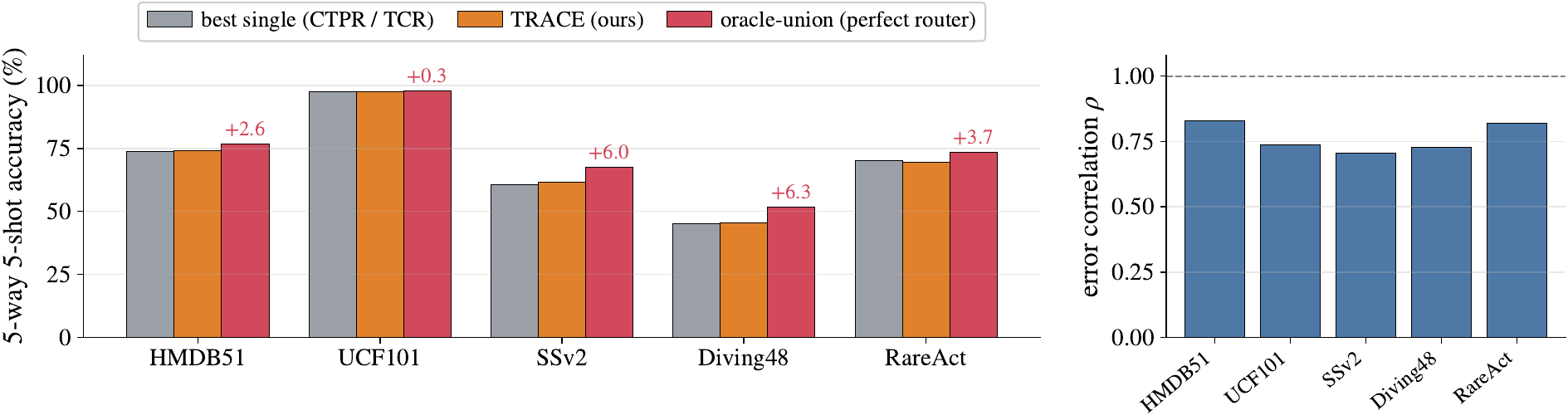}
\caption{\textbf{Complementarity without a reliable routing signal at 5-shot.}
\textbf{Left:} per dataset, the left bar shows the better of CTPR and TCR, the centre bar shows \sotd{}, and the right bar shows the oracle-routing ceiling. \textbf{Right:} per-episode error correlation $\rho$ between CTPR and TCR. Higher correlation indicates that the two estimators tend to fail on the same episodes, making reliable episode-level routing difficult.
}
\label{fig:comp5}
\end{figure}

Figure~\ref{fig:gain5} shows that most episodes continue to benefit from \sotd{}, indicating that the average improvement does not arise from a small number of outliers. Figure~\ref{fig:comp5} confirms that CTPR and TCR remain complementary, while leaving a sizeable oracle-routing gap that cannot be reliably exploited using the available label-free signals.

Together, Figures~\ref{fig:gain5} and~\ref{fig:comp5} show that the mechanism identified in the main paper remains unchanged at 5-shot: \sotd{} delivers broadly distributed improvements while benefiting from complementary prototype and boundary refinement.

\section{Failure Cases Follow the Estimation-Gap Framework}
\label{sec:supp-exceptions}

This section provides a cell-level verification of the two-factor explanation proposed in the main paper. The benefit of \sotd{} is governed by (i) the size of the estimation gap and (ii) the reliability of the confident pseudo-labels used for transductive refinement. The few settings in which \sotd{} is not the strongest method therefore provide a useful stress test of the estimation-gap framework.

Across three heads, five target datasets, and two shot settings ($30$ cells in total), \sotd{} is the strongest method in $25$ cells and not the strongest in $5$. As we show below, both the occurrence and the magnitude of these exceptions are consistent with the framework's predictions.

\paragraph{The gain tracks pseudo-label precision.}

Across all analyses, the same pattern emerges: \sotd{} benefits only when the confident pseudo-labels provide information beyond the baseline prediction. Table~\ref{tab:supp-prec} measures this relationship directly.

Within the Good-embed head (top block), the \sotd{} gain decreases monotonically with pseudo-label precision, from $+8.7$ points on UCF101, where confident pseudo-labels are $85\%$ correct, to $0$ on Diving48, where the base accuracy of $25.7\%$ is only slightly above the five-way chance level of $20\%$. In this regime, the pseudo-labels contain little usable information for transductive refinement.

The cross-head comparison on Diving48 (bottom block) provides the complementary perspective. TAMT and ProtoNet retain only around $20\%$ of queries, but those selected pseudo-labels achieve $43$--$45\%$ precision, substantially above the average episode accuracy. This is sufficient for \sotd{} to improve performance. In contrast, the Good-embed head is over-confident: the threshold retains almost all queries, causing the selected precision to collapse to the base accuracy. When the representation cannot produce a reliable confident set, there is little signal available for transductive refinement to exploit.

\begin{table}[h]
\centering\small
\input{tables/supp_precision}
\caption{\textbf{Pseudo-label quality explains \sotd{} gains (1-shot).}
\emph{Base acc.}: accuracy of the frozen base model. \emph{Conf.\ prec.}: precision of pseudo-labels above the confidence threshold ($\tau{=}0.5$). \emph{Coverage}: fraction of queries above the threshold. \emph{\sotd{} gain}: improvement over the plain baseline.}
\label{tab:supp-prec}
\end{table}

\paragraph{The remaining differences are small.}

Table~\ref{tab:supp-ci} reports the five non-winning cells together with $95\%$ confidence intervals computed over $10{,}000$ episodes.

Three of the five cases (ProtoNet and Good-embed on Diving48 at 1-shot, and TAMT on UCF101 at 5-shot) are statistically indistinguishable from the strongest component: the gap between \sotd{} and the best method falls within the confidence interval. The remaining two cases occur on RareAct and Diving48 at 5-shot, where TCR alone exceeds \sotd{} by a small but statistically significant margin ($0.59$ and $0.62$ points, respectively).

These cases occur precisely in the saturated 5-shot regime, where the estimation gap is already small and support prototypes are well estimated. Under these conditions, the refined-prototype initialisation used by \sotd{} contributes little beyond the support prototype itself, making the plain support initialisation of TCR occasionally marginally preferable. Importantly, \sotd{} still improves over the plain baseline in both cases and remains the strongest method on average across all evaluated heads.

\begin{table}[h]
\centering\small
\resizebox{\linewidth}{!}{\input{tables/supp_ci}}
\caption{\textbf{The remaining failure cases are either tied or marginal.}
Results with $95\%$ confidence intervals over $10{,}000$ episodes. $\Delta$ denotes the gap between the strongest component and \sotd{}.}
\label{tab:supp-ci}
\end{table}

\paragraph{Summary.}

The analysis throughout this paper identifies two quantities that govern the benefit of test-time refinement: the size of the estimation gap and the reliability of the confident pseudo-labels. The five non-winning cells arise precisely where one of these quantities becomes small. Either the estimation gap has largely disappeared (the saturated 5-shot regime), or the pseudo-labels are insufficiently reliable (a weak head on the most challenging target). In both situations, there is little signal available for transductive refinement to exploit.

These five cells do not contradict the estimation-gap framework; they are exactly the cases where it predicts little room for improvement.

\section{Backbone Generalisation Beyond VideoMAE}
\label{sec:supp-timesformer}

The gains of \sotd{} are not tied to a particular backbone architecture. To verify this directly, we evaluate the method unchanged on a TimeSformer backbone in place of the VideoMAE-S encoder used throughout the main paper. If the estimation-gap account is correct, similar improvements should emerge despite the change in encoder architecture.

We follow the identical protocol used throughout the paper: the same target datasets, Kinetics-400 source domain, evaluation splits, $10{,}000$ test episodes, and \sotd{} configuration ($\tau{=}0.5$, $T{=}3$, $\alpha{=}0.1$, $\beta{=}1.0$, $\gamma{=}15$, $\lambda{=}0.5$, and $100$ Adam steps). Table~\ref{tab:supp-timesformer} reports the results.

\begin{table}[h]
\centering\small
\begin{tabular}{llcccccc}
\toprule
Shot & Method & HMDB51 & UCF101 & SSv2 & Diving48 & RareAct & Avg. \\
\midrule
\multicolumn{8}{l}{\emph{TimeSformer backbone, Kinetics-400 source}}\\
1 & Baseline       & 56.72 & 84.41 & 40.28 & 31.15 & 45.44 & 51.60 \\
1 & \sotd{} (ours) & \textbf{63.65} & \textbf{90.13} & \textbf{42.80} & \textbf{32.67} & \textbf{50.51} & \textbf{55.95} \\
\midrule
5 & Baseline       & 70.13 & 94.17 & 57.61 & 42.99 & 68.35 & 66.65 \\
5 & \sotd{} (ours) & \textbf{71.97} & \textbf{97.02} & \textbf{59.84} & \textbf{44.03} & \textbf{70.06} & \textbf{68.58} \\
\bottomrule
\end{tabular}
\caption{\textbf{The gains of \sotd{} transfer across backbone architectures.}
5-way accuracy (\%) under the Kinetics-400 source setting. Only the frozen backbone changes from VideoMAE-S to TimeSformer; all \sotd{} hyperparameters remain unchanged. Best results per column are highlighted in \textbf{bold}.}
\label{tab:supp-timesformer}
\end{table}

\paragraph{Takeaway.}
\sotd{} transfers across backbone architectures. On the TimeSformer base, it improves the five-target average by $+4.35$ points at 1-shot and $+1.93$ points at 5-shot, outperforming the baseline on every dataset and shot.

The pattern closely mirrors that observed with VideoMAE-S in the main paper. Gains are largest at 1-shot, where the estimation gap is largest, and smaller at 5-shot, where the gap contracts. Improvements also remain smallest on Diving48, the most challenging temporally-dependent target, and larger on the appearance-driven targets.

Despite starting from a weaker baseline than VideoMAE-S, TimeSformer receives a comparable absolute improvement ($+4.35$ vs.\ $+4.98$ at 1-shot; $+1.93$ vs.\ $+1.50$ at 5-shot). This suggests that \sotd{} is not exploiting properties of a specific encoder. Instead, it acts on estimation errors induced by support scarcity, supporting the central claim that the estimation-gap framework describes a property of cross-domain few-shot inference rather than of any particular backbone.

\section{Simple Transductive Transforms Do Not Explain the Gains}
\label{sec:supp-ptmap}

An alternative explanation for the gains of \sotd{} is that they arise from generic transductive feature preprocessing rather than from estimation-gap recovery. To test this, we compare against two representative operators applied directly to the same frozen TAMT embeddings used throughout the paper: (i) feature centring followed by L2 normalisation, and (ii) the power-transform plus L2 preprocessing used in PT-MAP. Both are applied before CTPR under the identical evaluation protocol. Table~\ref{tab:supp-ptmap} reports the results.

\begin{table}[h]
\centering\small
\resizebox{\linewidth}{!}{%
\begin{tabular}{lcccccc}
\toprule
Method (5-way 1-shot) & HMDB51 & UCF101 & SSv2 & Diving48 & RareAct & Avg. \\
\midrule
TAMT plain (baseline) & 58.19 & 86.04 & 41.14 & 32.36 & 46.33 & 52.81 \\
\midrule
Feature centring + L2-norm + CTPR & 60.30 & 87.91 & 41.66 & 34.06 & 48.14 & 54.41 \\
Power-transform$+$L2 (PT-MAP preproc.) + CTPR & 58.71 & 86.17 & 40.29 & 31.53 & 46.13 & 52.57 \\
\midrule
\sotd{} (ours) & \textbf{64.81} & \textbf{95.35} & \textbf{43.56} & \textbf{34.15} & \textbf{51.09} & \textbf{57.79} \\
\bottomrule
\end{tabular}%
}
\caption{\textbf{\sotd{} substantially outperforms alternative transductive feature transforms.}
5-way 1-shot accuracy (\%). All methods operate on the same frozen TAMT embeddings. Feature preprocessing is applied independently within each episode before CTPR. Best results per column are highlighted in \textbf{bold}.}
\label{tab:supp-ptmap}
\end{table}

\paragraph{Takeaway.}

Simple feature-space transforms account for only a small fraction of the gains achieved by \sotd{}. Feature centring with L2 normalisation improves the five-dataset average by $+1.60$ points over the plain baseline, whereas PT-MAP preprocessing is nearly inert and performs slightly below the baseline on average. Both remain substantially below \sotd{}, which improves the same baseline by $+4.98$ points.

This behaviour is consistent with the design of the TAMT representation. The moment-tuning head already models and removes channel-wise moments, producing embeddings that are comparatively well-conditioned. As a result, the heavy-tail correction that underlies PT-MAP contributes little additional benefit in this setting, while simple feature normalisation provides only modest gains.

In contrast, \sotd{} does not seek to improve the feature representation itself. Instead, it targets the estimation errors identified by the oracle analysis through prototype refinement (CTPR) and classifier refinement (TCR). The substantially larger gains therefore suggest that the remaining headroom in TAMT arises primarily from support-limited estimation rather than from feature conditioning. This provides further evidence for the central claim of the paper: under strong frozen representations, estimation---not feature normalisation---becomes the dominant bottleneck in CD-FSAR.

\section{Transductive Baselines, Class Imbalance, and Query Budget}
\label{sec:supp-robustness}

This section examines three questions about the evaluation protocol: how
\sotd{} compares with established transductive methods under identical
conditions, whether the gains survive class-imbalanced query sets, and how
they depend on the query budget $M$. All experiments use the frozen TAMT
embeddings and the evaluation episodes of the main paper.

\paragraph{Transductive baselines on identical features.}
Table~\ref{tab:supp-transductive} compares \sotd{} with BD-CSPN (Liu
\etal, ECCV 2020), LaplacianShot (Ziko \etal, ICML 2020), a three-round
balanced Sinkhorn assignment loop in the style of PT-MAP, and TIM (Boudiaf
\etal, NeurIPS 2020). Every method receives the same frozen features, the
same $10{,}000$ episodes per dataset, and the same query batches; all are
training-free. \sotd{} attains the best average and the best accuracy
on four of the five datasets; on SSv2 it is statistically tied with the
Sinkhorn variant under paired $95\%$ confidence intervals. Transductive
FSAR methods that couple transduction to a meta-trained alignment
architecture and a single-domain protocol (for example Nguyen \etal,
ECCV 2022) cannot run on frozen features, so the table covers the
operators that transfer to this setting.

\begin{table}[h]
\centering\small
\resizebox{\linewidth}{!}{%
\begin{tabular}{lcccccc}
\toprule
Method (5-way 1-shot) & HMDB51 & UCF101 & SSv2 & Diving48 & RareAct & Avg. \\
\midrule
TAMT plain (baseline) & 58.19 & 86.04 & 41.14 & 32.36 & 46.33 & 52.81 \\
LaplacianShot & 58.39 & 88.92 & 40.49 & 30.87 & 46.64 & 53.06 \\
BD-CSPN & 59.74 & 90.06 & 41.85 & 33.07 & 48.08 & 54.56 \\
Balanced Sinkhorn & 62.09 & 94.64 & \textbf{43.93} & 33.10 & 49.27 & 56.61 \\
TIM & 63.50 & 93.44 & 43.12 & 33.96 & 50.02 & 56.81 \\
\midrule
CTPR (ours) & 61.82 & 92.73 & 42.74 & 32.74 & 49.12 & 55.83 \\
\sotd{} (ours) & \textbf{64.81} & \textbf{95.35} & 43.56 & \textbf{34.15} & \textbf{51.09} & \textbf{57.79} \\
\bottomrule
\end{tabular}%
}
\caption{\textbf{Transductive methods on identical frozen features and
episodes.} 5-way 1-shot accuracy (\%; $10{,}000$ episodes per dataset);
all methods are training-free and receive the same query batches. On
SSv2, \sotd{} and the balanced Sinkhorn variant are statistically tied
under paired $95\%$ confidence intervals. Best per column in
\textbf{bold}.}
\label{tab:supp-transductive}
\end{table}

\begin{figure}[h]
\centering
\includegraphics[width=\linewidth]{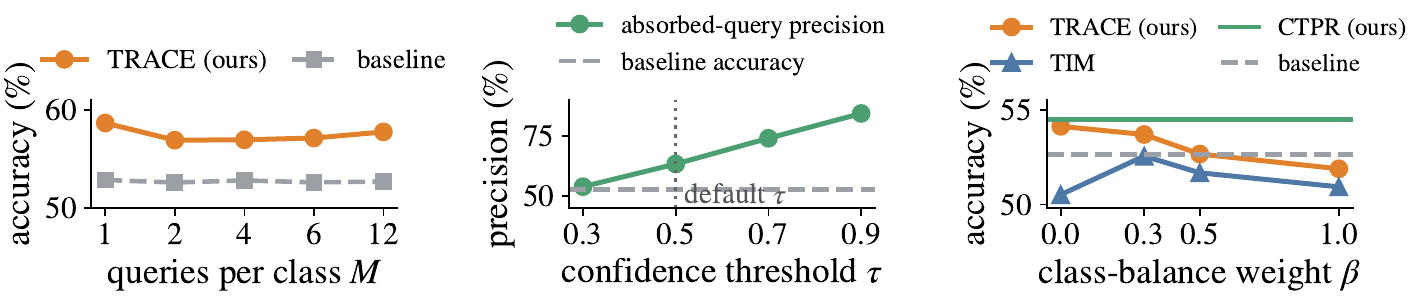}
\caption{\textbf{Robustness studies} (5-way 1-shot, averages over the
five target datasets). Left: query-budget sweep; \sotd{} keeps its full
margin at every $M$. Middle: precision of CTPR-absorbed queries exceeds
baseline accuracy and rises with $\tau$. Right: $\beta$ sweep under
Dirichlet-imbalanced queries; \sotd{} stays above TIM at every matched
$\beta$.}
\label{fig:supp-robustness}
\end{figure}

\paragraph{Class-imbalanced query sets.}
TCR encourages class-balanced predictions, so we evaluate it where that
prior is violated. Following the protocol of Veilleux \etal\ (NeurIPS
2021), each episode draws its query counts from a Dirichlet distribution
($\alpha_{\mathrm{Dir}}{=}2$, $30$ queries per episode, $2{,}000$ episodes
per dataset). CTPR makes no balance assumption and stays positive on all
five datasets ($+1.8$ points on average, significant under paired
confidence intervals). The balance prior enters only through the
marginal-entropy weight $\beta$ of the TCR objective: accuracy rises
monotonically as $\beta$ decreases, and $\beta\!\le\!0.3$ keeps \sotd{}
above the baseline ($+1.0$). At every matched $\beta$, \sotd{} remains
above TIM on every dataset, and the margin grows as the prior is removed
($+1.1$ at $\beta{=}0.3$, $+3.6$ at $\beta{=}0$;
Fig.~\ref{fig:supp-robustness}, right). The robustness therefore comes
from the composition with CTPR initialisation rather than from the
entropy objective alone; under unknown class priors we recommend a low
$\beta$.

\paragraph{Query budget and the confidence assumption.}
Varying the number of queries per class, $M\in\{1,\dots,12\}$, leaves the
margin over the baseline intact ($+5.8$ at $M{=}1$, $+5.1$ at $M{=}12$;
Fig.~\ref{fig:supp-robustness}, left); $M{=}12$ follows the TAMT
evaluation protocol. The confidence assumption behind CTPR also holds in
direct measurement: at $\tau{=}0.5$ the precision of the absorbed queries
exceeds the baseline accuracy on every dataset ($61.3$ vs.\ $41.1$ on
SSv2; $63.4$ vs.\ $52.8$ on average) and rises with $\tau$. Absorption
also self-limits where pseudo-labels are weak: $17\%$ of queries are
absorbed on Diving48 against $96\%$ on UCF101
(Fig.~\ref{fig:supp-robustness}, middle).

%% file: tables/gentable_T4_1shot.tex
\begin{tabular}{llcccccc}
\toprule
Base & Method & HMDB51 & UCF101 & SSv2 & Diving48 & RareAct & Avg. \\
\midrule
ProtoNet & Baseline & 58.26 & 87.63 & 39.93 & 31.89 & 47.86 & 53.11 \\
 & +CTPR & 62.30 & 93.43 & 41.44 & \textbf{32.37} & 51.06 & 56.12 \\
 & +TCR & 63.10 & 94.27 & 41.98 & 31.57 & 51.12 & 56.41 \\
 & TRACE (ours) & \textbf{64.87} & \textbf{95.95} & \textbf{42.54} & 32.36 & \textbf{52.29} & \textbf{57.60} \\
\midrule
Good-embed & Baseline & 51.94 & 84.75 & 31.08 & 25.70 & 47.26 & 48.15 \\
 & +CTPR & 55.42 & 91.46 & 31.42 & 25.49 & 50.27 & 50.81 \\
 & +TCR & 54.06 & 89.66 & 31.30 & \textbf{25.82} & 48.54 & 49.88 \\
 & TRACE (ours) & \textbf{56.76} & \textbf{93.42} & \textbf{31.60} & 25.69 & \textbf{50.83} & \textbf{51.66} \\
\bottomrule
\end{tabular}

%% file: tables/gentable_T4_5shot.tex
\begin{tabular}{llcccccc}
\toprule
Base & Method & HMDB51 & UCF101 & SSv2 & Diving48 & RareAct & Avg. \\
\midrule
ProtoNet & Baseline & 73.59 & 96.04 & 55.81 & 45.14 & 67.70 & 67.66 \\
 & +CTPR & 75.18 & 96.55 & 56.75 & 45.55 & 68.81 & 68.57 \\
 & +TCR & 76.40 & \textbf{97.86} & 57.39 & 45.38 & 68.20 & 69.05 \\
 & TRACE (ours) & \textbf{76.92} & \textbf{97.86} & \textbf{58.03} & \textbf{45.74} & \textbf{68.97} & \textbf{69.50} \\
\midrule
Good-embed & Baseline & 68.73 & 95.47 & 42.18 & 31.99 & 67.59 & 61.19 \\
 & +CTPR & 69.64 & 96.01 & 42.49 & 31.65 & 68.45 & 61.65 \\
 & +TCR & 70.94 & 97.04 & 43.71 & \textbf{32.74} & 70.72 & 63.03 \\
 & TRACE (ours) & \textbf{71.73} & \textbf{97.29} & \textbf{43.79} & 32.12 & \textbf{70.89} & \textbf{63.16} \\
\bottomrule
\end{tabular}

%% file: tables/supp_precision.tex
\begin{tabular}{llcccc}
\toprule
& Setting & Base acc. & Conf.\ prec. & Coverage & \sotd{} gain \\
\midrule
\multicolumn{6}{l}{\emph{Good-embedding head, 1-shot (vary target)}}\\
& UCF101 & 84.8 & 84.8 & 100.0 & $+8.67$ \\
& HMDB51 & 51.9 & 52.0 & 99.9 & $+4.82$ \\
& RareAct & 47.3 & 47.3 & 99.9 & $+3.57$ \\
& SSv2 & 31.1 & 31.1 & 99.9 & $+0.52$ \\
& Diving48 & 25.7 & 25.7 & 99.8 & $-0.01$ \\
\midrule
\multicolumn{6}{l}{\emph{Diving48, 1-shot (vary base, hardest target)}}\\
& TAMT & 32.4 & 45.0 & 17.7 & $+1.79$ \\
& ProtoNet & 31.9 & 43.2 & 20.5 & $+0.47$ \\
& Good-embed & 25.7 & 25.7 & 99.8 & $-0.01$ \\
\bottomrule
\end{tabular}

%% file: tables/supp_ci.tex
\begin{tabular}{llccccc}
\toprule
Shot & Head / target & plain & +CTPR & +TCR & \sotd{} & $\Delta$(best$-$\sotd{}) \\
\midrule
1 & ProtoNet / Diving48 & $31.89{\scriptstyle\,\pm0.15}$ & $32.37{\scriptstyle\,\pm0.16}$ & $31.57{\scriptstyle\,\pm0.15}$ & $32.36{\scriptstyle\,\pm0.15}$ & $0.01$ \\
1 & Good-embed / Diving48 & $25.70{\scriptstyle\,\pm0.12}$ & $25.49{\scriptstyle\,\pm0.13}$ & $25.82{\scriptstyle\,\pm0.13}$ & $25.69{\scriptstyle\,\pm0.13}$ & $0.13$ \\
5 & TAMT / UCF101 & $95.96{\scriptstyle\,\pm0.06}$ & $96.48{\scriptstyle\,\pm0.06}$ & $97.47{\scriptstyle\,\pm0.05}$ & $97.43{\scriptstyle\,\pm0.05}$ & $0.04$ \\
5 & TAMT / RareAct & $68.18{\scriptstyle\,\pm0.15}$ & $68.90{\scriptstyle\,\pm0.16}$ & $70.26{\scriptstyle\,\pm0.15}$ & $69.67{\scriptstyle\,\pm0.16}$ & $0.59$ \\
5 & Good-embed / Diving48 & $31.99{\scriptstyle\,\pm0.13}$ & $31.65{\scriptstyle\,\pm0.14}$ & $32.74{\scriptstyle\,\pm0.13}$ & $32.12{\scriptstyle\,\pm0.14}$ & $0.62$ \\
\bottomrule
\end{tabular}